\documentclass{article}

\usepackage[preprint]{neurips_2024}

\usepackage[utf8]{inputenc} % allow utf-8 input
\usepackage[T1]{fontenc}    % use 8-bit T1 fonts
\usepackage{amsfonts}       % blackboard math symbols
\usepackage{algorithm}
\usepackage{algpseudocode}
\usepackage{amsmath}
\usepackage[pdftex]{graphicx}
\usepackage{nicefrac}       % compact symbols for 1/2, etc.
\usepackage{microtype}      % microtypography
\usepackage[table,xcdraw]{xcolor} % for coloring cells
\usepackage{hhline} % for better lines in the table
\usepackage{caption}
\usepackage{wrapfig}
\usepackage{hyperref}

\title{Greedy Output Approximation: Towards Efficient Structured Pruning for LLMs Without Retraining}

\author{ Jianwei Li\thanks{Department of Computer Science at North Carolina State University, Email: \texttt{jli265@ncsu.edu}} \\
	\and
        Yijun Dong\thanks{Courant Institute of Mathematical Sciences at New York University, Email: \texttt{yd1319@nyu.edu}} \\
	\and
	  Qi Lei\thanks{Center for Data Science at New York University, Email: \texttt{ql518@nyu.edu}}\\ %\thanks{Corresponding author} 
}

\begin{document}

\maketitle

\begin{abstract}

%The unprecedented recent successes of large language models (LLMs) are built upon their exponentially growing model size. The enormous cost of training, finetuning, and even inferencing models with billions of parameters brings up the importance of network pruning---reducing model parameters while preserving the performance. However, the astronomical scales of LLMs render traditional pruning methods that involve iterative pruning and retraining less effective due to the prohibitive computational cost. While single-shot network pruning methods are getting increasing attention as an affordable and competitive alternative, these methods still require at least one round of retraining after pruning, with their performance relying heavily on the diversity of datasets.

%To address these challenges, this paper simplifies the pruning process of Transformer-based LLMs by successfully identifying a depth-2 pruning structure that operates independently. Furthermore, we introduce two inference-aware pruning criteria derived from the optimization perspective of output approximation, which surpass traditional training-aware metrics such as gradient and Hessian. Moreover, we introduce a two-step reconstruction technique to mitigate pruning errors without model retraining. Our experimental results showcase the superior performance of this approach across various datasets and models, markedly reducing both computational costs and hardware requirements. 

To remove redundant components of large language models (LLMs) without incurring significant computational costs, this work focuses on single-shot pruning without a retraining phase. We simplify the pruning process for Transformer-based LLMs by identifying a depth-2 pruning structure that functions independently. Additionally, we propose two inference-aware pruning criteria derived from the optimization perspective of output approximation, which outperforms traditional training-aware metrics such as gradient and Hessian. We also introduce a two-step reconstruction technique to mitigate pruning errors without model retraining. Experimental results demonstrate that our approach significantly reduces computational costs and hardware requirements while maintaining superior performance across various datasets and models.
\end{abstract}

\section{Introduction}

With the development of LLMs displaying emergent capabilities like sophisticated reasoning, the focus of the community has shifted to models with billions of parameters, for example, GPT-4 and Llama2~\cite{achiam2023gpt,touvron2023llama2}. This transition introduces unprecedented computational costs both in the training and the inference phases~\cite{touvron2023llama,le2023bloom,team2023gemini,banks2024gemma}. To address this challenge, pruning plays a constructive role by removing redundant components from models, thereby reducing computational costs~\cite{Gordon:2020,Prasanna:2020,Wang:2020,li2023breaking}. Notably, designing an optimal pruning strategy is an NP-hard problem (as it reduces to subset selection) and requires balancing accuracy, sparsity, generalizability, pruning costs, and hardware compatibility in practice~\cite{xu-etal-2021-beyond,li2023towards,du-etal-2023-robustness}. Traditional pruning methods primarily focus on accuracy and sparsity, often neglecting other key factors. They typically involve model retraining and knowledge distillation to mitigate pruning errors. However, with current LLMs featuring billions of parameters, the training process is already a significant challenge, making the additional cost of model retraining even more unaffordable~\cite{frantar2022optimal,frantar2023sparsegpt}. Given these challenges, there's a pressing need for more efficient pruning approaches.

Recently, some works have focused on structured pruning on pre-trained LLMs, directly addressing hardware compatibility and generalizability. This approach allows them to concentrate on the remaining trade-off factors: sparsity, accuracy, and pruning cost. For example, methods like LLM-Pruner, Shortened LLaMA, and Sheared LLaMA use a single-shot pruning strategy that requires only one round of retraining~\cite{kim2024shortened,xia2023sheared,ma2023llm}. On the other hand, strategies such as FLAP, OPTIN, Sliced GPT, LLM Surgeon, Wanda, ZipLM, and KRP seek to eliminate the need for model retraining entirely~\cite{an2024fluctuation,ashkboos2024slicegpt,van2023llm,sun2023simple,kurtic2024ziplm,khaki2024need,li2023towards}. However, these approaches have respective limitations, such as high computational costs from the calculation of higher-order information, a lack of fully structured pruning patterns~\cite{pool2021accelerating}, or compromised performance in some cases. The development of these methods marks a crucial phase in the evolution of LLMs, aiming to enhance model capabilities while ensuring computational efficiency.

\begin{figure}[!t]
    \center
    \includegraphics[width=\textwidth]{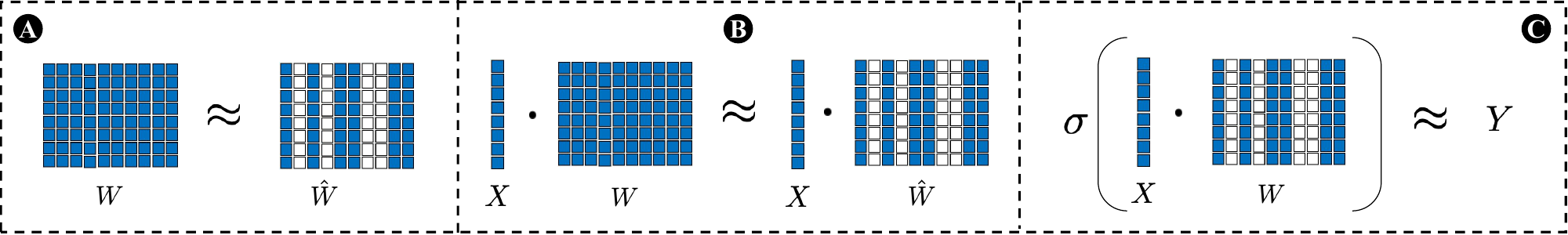}
    \caption{Pruning metric analysis from the optimization perspective \textbf{A:} Function Approximation; \textbf{B:} Output Approximation; \textbf{C:} Objective Approximation.}
    \label{fig:1}
    \vspace{-0.1cm}
\end{figure}

With the existing challenges, we call for a pruning strategy that better trades off the accuracy, structure preservation, and computational costs. We delve into this kind of strategy by answering the following essential questions: 

\textbf{Question~1.} \textit{Does an inherent \textbf{pruning structure} exist in Transformer-based language models?}

We discovered depth-2 pruning modules within Transformer architecture by analyzing structured pruning from both input and output channels. These structures preserve feature knowledge while reducing pruning complexity from residual connections.

\textbf{Question~2.} \textit{Is there an effective \textbf{pruning criterion} that does not require training awareness?}

We identified two efficient and high-performing inference-aware pruning metrics based on output approximation for Transformer models. They are simpler and more computationally efficient than training-aware metrics.

\textbf{Question~3.} \textit{Is there a low-computation \textbf{performance recovery} technique available?} 

Inspired by layer-wise reconstruction~\cite{li2023towards}, we developed a two-step module reconstruction strategy. This method updates the weights of the depth-2 module without computing parameter gradients, effectively minimizing pruning errors.

Answering the above questions altogether, this paper proposes an integrated and efficient pruning strategy with a focus on Transformer-based LLMs% and examines different pruning criteria from an optimization perspective
~\cite{vaswani2017attention}. Specifically, we categorize all pruning metrics into three groups based on their implicit purpose: function (weights) approximation, output approximation, and objective approximation. Following the output approximation route, we introduce a similarity-based pruning strategy that exploits the redundancy in multi-head attention mechanisms by removing heads that extract similar information first rather than those with minimal impact. Additionally, we propose a second-moment-based pruning approach also under the output approximation category, which stands out for its ability to integrate information across multiple layers. We apply this metric for both attention and feed-forward modules to remove the elements with minimal impact on the model's performance. Finally, we develop an optimization technique that eliminates the need for higher-order information by greedily reducing pruning error through weight reconstruction of the subsequent dense module. Our structured pruning experiments on pre-trained LLMs ranging from millions to billions of parameters demonstrate that our method ensures generalizability, hardware compatibility, and minimal pruning cost. Moreover, it outperforms or achieves comparative performance to other non-retraining methods and even some methods that require retraining.

\section{Related Work\label{sec:related}}

\textbf{Pruning and Structured Pruning:} Pruning is a technique used in machine learning to reduce the size of a model by eliminating unnecessary parameters, which can lead to reductions in storage requirements and computational complexity without significantly affecting the model's performance~\cite{frantar2022optimal,Frankle:2020}. This process involves identifying and removing the parts of the model that have the least impact on its output, such as weights in a neural network with small magnitudes. By doing so, pruning discovers a more efficient model that is faster to execute and easier to deploy on devices with limited resources. Structured pruning, a method that imposes more constraints, focuses on eliminating entire units or structures within the model, such as neurons, channels, or layers, instead of individual weights~\cite{anwar2017structured,fang2023depgraph}. Being the focus of our paper, structured pruning is especially advantageous due to its compatibility with standard hardware, whereas unstructured pruning requires specially designed accelerators to deploy in practical scenarios.

\textbf{Data-free/dependent and Training/Inference-aware Metrics:} When choosing which redundant components to remove from a model, the selection is typically guided by specific metrics~\cite{hoefler2021sparsity}. These metrics can be broadly divided into data-free and data-dependent categories, depending on whether they rely on specific datasets. Additionally, they can be categorized as training-aware or inference-aware, based on whether they require model backpropagation. This paper focuses on inference-aware metrics and explores both data-free and data-dependent versions.

\textbf{Efficient and Low-Resource Pruning:} As the number of parameters in LLMs increases, the quest for efficient pruning has become paramount. Methods such as LLM-Pruner, Sheared LLaMA, and Shortened LLaMA adopt a single-shot pruning strategy~\cite{kim2024shortened, xia2023sheared, ma2023llm}. Yet, these approaches depend on computationally expensive metrics and still require retraining to minimize pruning-induced errors. In contrast, methods like OPTIN, Sliced GPT, LLM Surgeo, ZipLM, and KRP eliminate the need for retraining but still rely heavily on computationally expensive second-order Hessian information, which is a significant drawback for large-dimensional models~\cite{ashkboos2024slicegpt, sun2023simple, kurtic2024ziplm, khaki2024need, li2023towards}. Meanwhile, FLAP and Wanda design specific pruning metrics that share similar ideas with the methods from the pre-deep learning age~\cite{engelbrecht1999variance,sietsma1988neural,engelbrecht1996sensitivity,thimm1995evaluating}, significantly reducing computational demand~\cite{an2024fluctuation, van2023llm}. This paper proposes a method that eliminates the need for both model retraining and computationally expensive metrics, offering superior or comparative performance compared to other non-retraining methods and even some methods that require retraining.

\section{Methodology\label{section:5}}

This section outlines our structured pruning scheme, which consists of three key components: pruning structure recognition, pruning criteria definition, and post-pruning recovery strategy.

% \textbf{Notations:} To better demonstrate our method, let us first establish the notations. We focus on the pruning of Transformer-based large language models, thus we refer to the attention mechanism as $\mathbf{Cat}_{i=1}^{h}[\sigma_{1}(\mathbf{X}\mathbf{W}_{i}^{K}\mathbf{W}_{i}^{Q}\mathbf{X}^\top)\mathbf{X}\mathbf{W}_{i}^{V}]\mathbf{W}^{O}$, with $i$ indicating the attention head index. The symbols $W^{K}$, $W^{Q}$, $W^{V}$ and $W^{O}$ represent the weights for the key, query, value, and output in the attention block, respectively. For the general and gated feed-forward module, we denote the logic as $\mathbf{W}^{D}\sigma_2(\mathbf{W}^{U}\mathbf{X})$ and $\mathbf{W}^{D}(\mathbf{W}^{U}\mathbf{X} * \sigma_2(\mathbf{W}^{G}\mathbf{X}))$. Here, $W^{U}$, $W^{D}$, and $W^{G}$ stand for the weights for upward projection, downward projection, and gate projection. $\sigma$ refers to the activation function for all of them, which can be SoftMax, ReLU, GeLU, or SiLU function. 

\subsection{Pruning Structure Recognition\label{sec:structure}}

Our approach involves single-shot pruning and targets structured components, such as entire rows or columns of weight matrices, rather than individual weights. We do not discuss layer or block pruning, as it disrupts inherent model correlations and requires retraining to restore layer dependencies.

\subsubsection{Input or Output Channel Pruning for Sequential Layers\label{sec:in-out}}
To clarify structured pruning, it is important to understand that pruning neurons can be approached in two directions: input channels and output channels. Consider a linear function \( f(X) = XW \), where \( X \in \mathbb{R}^{d_{in}} \) is the input and \( W \in \mathbb{R}^{d_{in} \times d_{out}} \) is the weight matrix. When we prune neurons, we typically refer to pruning the output channels of \( W \), since the number of neurons generally corresponds to the number of output channels in each layer. After pruning, the weight matrix becomes \( \hat{W} \in \mathbb{R}^{d_{in} \times d'_{out}} \), where \( d'_{out} < d_{out} \). Alternatively, pruning the input channels of \( W \) equates to pruning the input \( X \), also known as feature selection. This paper focuses on a static approach to feature selection, where the same channels are removed for all samples, making feature selection equivalent to output channel pruning in the previous layer. An interesting phenomenon arises: in depth-2 sequential linear layers, pruning the input channels of the second layer simultaneously pruning the output channels of the first layer, using identical pruning indices. In contrast, pruning the output channels of the second layer does not affect the first layer. Both input and output channel pruning contribute to model compression, but they may have different impacts on performance.

\begin{figure*}[!htb]
    \center
    \includegraphics[width=\textwidth]{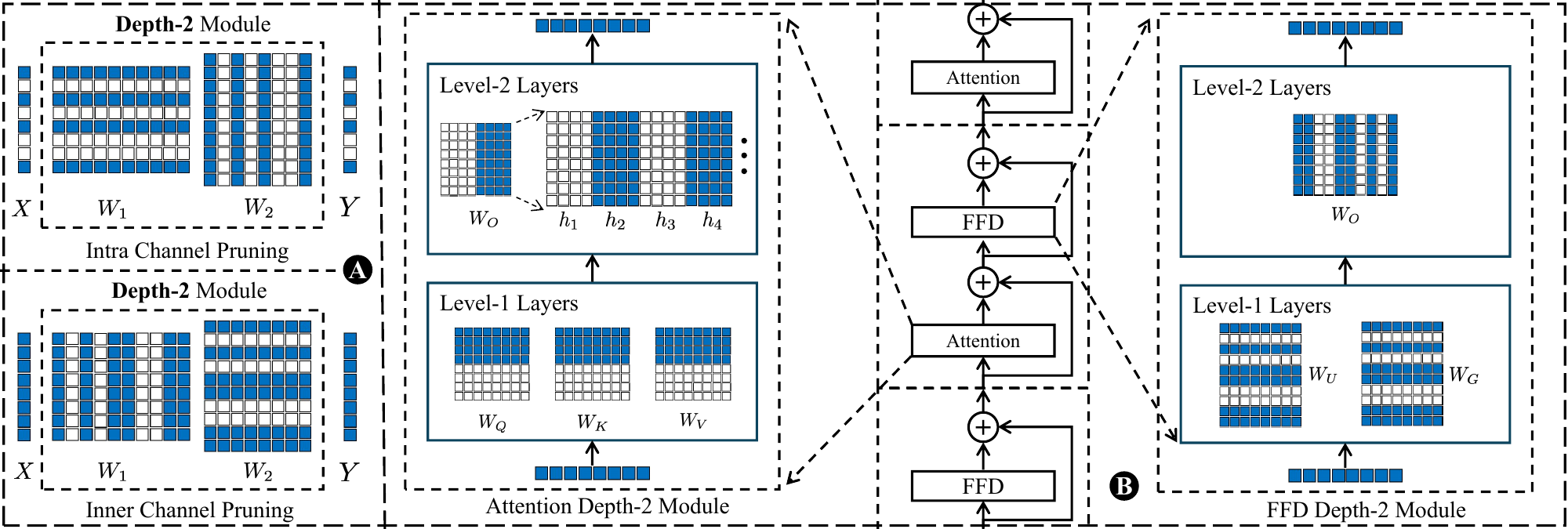}
    \caption{Pruning structure recognition.~\textbf{A}: Two pruning strategies for the depth-2 module.~\textbf{B}: Depth-2 modules identification in Transformer-based LLMs.}
    \label{fig:2}
    % \vspace{-0.1cm}
\end{figure*}

\subsubsection{Input or Output Channel Pruning for Transformer \label{sec:two-structure}}
\textbf{Depth-2 Module Identification:} In Transformer-based language models, the attention and feed-forward modules function as sequential layers with a depth of 2. For the attention module, the first level includes the weight matrices \( W_Q \), \( W_K \), and \( W_V \), which operate in parallel. Pruning the input channels of any one of these matrices does not affect the output channels of the others. The second level consists of the weight matrix \( W_O \). The symbols \( W_Q \), \( W_K \), \( W_V \), and \( W_O \) represent the weights for the query, key, value, and output in the attention block, respectively. The feed-forward module follows a similar structure: the upward projection and gated projection occur at the first level, while the downward projection occurs at the second level. These depth-2 modules have a unique characteristic: when pruning the input channels of layers at the second level, the output channel indices of the first-level layers must correspondingly match. This ensures that the structural integrity of the model is maintained while pruning.

\textbf{Pruning Strategies for Depth-2 Modules:} Given these depth-2 modules, we can employ two pruning strategies to achieve the same compression ratio. The first strategy involves pruning the output channels of the layers in the first level while concurrently pruning the input channels of the layers in the second level. This approach ensures that the dependencies outside the module remain invariant. The second strategy involves pruning the output channels and the initial input \(X\) to the entire depth-2 module. In the context of the Transformer architecture, which consists of multiple such modules in sequence, pruning the initial input \(X\) is effectively equivalent to pruning the output channels of a preceding module in the sequence. As this dependency propagates backward through the layers, it ultimately affects the model's embedding layer, meaning we are directly pruning the channels of the token embeddings.

\textbf{Challenge of Residual Connection}: Without considering the loss of tokens' semantic information, the two pruning strategies described above should not differ significantly. However, residual connections impose substantial constraints on the second strategy. In the Transformer architecture, every depth-2 module determines residual connections. This means that the pruned channels must be strictly aligned across all modules. If the pruned indices of one of them do not align with others, it could lead to an unpredictable loss of information. This constraint severely limits the choice of channels for pruning and could significantly decrease performance. In contrast, the first strategy maintains a fixed number of output channels across these modules, avoiding this limitation. Each module can independently select which internal channels to prune based on its needs, resulting in a larger search space for optimization. This paper will focus on the first pruning strategy.

\textbf{Additional Structure for Attention Mechanism}: The intricate topology of the attention block introduces an additional constraint: pruning must be conducted at the level of entire heads, encompassing continuous portions of the channels. Fortunately, given the design philosophy of multi-head attention—that each head is designed to capture correlations between tokens independently—this setup easily leads to redundancy, making it highly amenable to similarity analysis. 

% Additionally, we highlight an alternative structure that has received less attention: pruning individual channels within each head, rather than removing entire heads. This strategy can serve as an alternative to full-head pruning or be employed in conjunction with it, offering further channel reduction. We describe more about this in Appendix X.

\subsection{Pruning Criteria Selection\label{sec:criteria}}

This section begins by examining different pruning criteria from an optimization perspective. Then, we introduce two specific pruning metrics for the aforementioned depth-2 modules. Finally, an intuitive magnitude-based pruning method is employed to remove the least important channels.

\subsubsection{Pruning Metric Analysis from an Optimization Perspective} 

Previous work has categorized pruning metrics based on their relationship with data, as discussed in Section~\ref{sec:related}. Diverging from these approaches, we analyze these metrics from an optimization perspective and describe in Fig.~\ref{fig:1}. Specifically, for a linear operation \( f(X) = XW \), our goal is to prune \( W \) while preserving the accuracy \( f(X) \approx Y \). To minimize pruning error, we can approximate \( W \), \( f(X) \), and \((f(X), Y)\). We term these strategies as function approximation, output approximation, and objective approximation, respectively. Function approximation focuses on directly approximating \( W \), which is equivalent to approximating the function itself. Typical metrics in this category include the L1 and L2 norms of weights or neurons. Output approximation seeks to approximate the result of \( XW \). Known metrics in this category include contribution energy, the sensitivity of \( f(X) \) to deviations in \( X \), and the variance or similarity score of \( f(X) \). Objective approximation aims to directly approximate accuracy. This category encompasses metrics such as first-order or second-order information and regularization scores. However, this type of metric is computationally expensive as the optimization process involves backward propagation and calculation of the Hessian Matrix. By analyzing these strategies, this paper proposes two new metrics to guide the pruning of LLMs.
\subsubsection{Similarity-based Metric for Attention Block~\label{sec:similarity}}

\begin{wrapfigure}{r}{0.40\textwidth}
  \centering
  \vspace{-0.4cm}
  \includegraphics[width=0.40\textwidth]{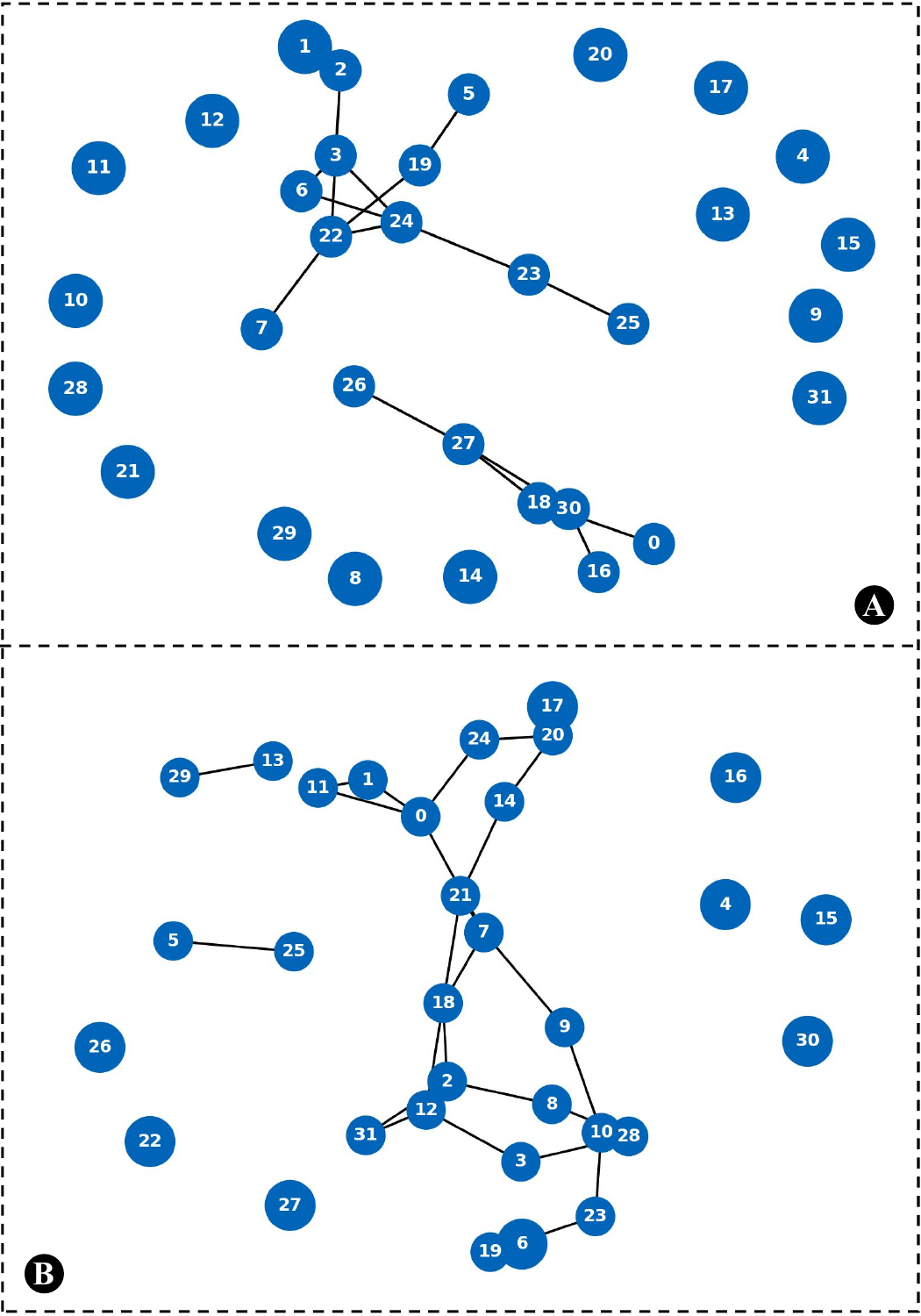}
  \captionof{figure}{Similarity visualization of attention heads in \textbf{A}: block 4 and \textbf{B}: block 5 for Llama-7B. Heads with divergence less than $\tau = 0.20$ are connected.}
  \label{fig:3}
  \vspace{-0.6cm}
\end{wrapfigure}

Previous research on pruning attention heads typically involves removing heads with the lowest importance scores. Surprisingly, our experiments indicate that random pruning also yields competitive results compared to magnitude-based pruning, especially when the pruning ratio is below 50\%. Further experimentation with different random seeds, leading to various head indices for pruning, consistently produces comparable results. Notably, nearly all heads have been selected for removal at some point during this process, suggesting a potential oversight in our initial understanding. Recall that different attention heads are intended to independently capture correlations between tokens. Thus, it's common for similar information to be extracted across different heads. This observation prompted us to reconsider our strategy: we prioritize removing similar heads before eliminating those with the least importance score. By identifying and pruning heads that capture redundant information, we can optimize the model more effectively while preserving its performance.

Previous studies have conducted similarity analysis between neurons~\cite{engelbrecht1996sensitivity,sietsma1988neural,engelbrecht1999variance}, examining the output differences across multiple samples to identify similar components. The redundant neurons are then removed, and the remaining neurons scale their weights or biases to minimize the impact of this removal. However, these methods are primarily effective in smaller neural networks, as the scaling technique struggles to handle the accumulated error across numerous layers. Fortunately, due to the parallelism and independence of attention heads, removing redundant heads does not lead to significant information loss that affects subsequent layers, thus eliminating the need for costly remedial operations. Based on this observation, we define a pairwise head divergence matrix \(D \in \mathbb{R}^{h \times h}\) for each attention module, where \(h\) refers to the number of heads. Specifically, given an attention score matrix \(Attn \in \mathbb{R}^{N \times h \times s \times s}\), where \(N\) represents the number of samples and \(s\) is the sequence length, let \(P_{i} \in \mathbb{R}^{N \times s \times s}\) and \(Q_{j} \in \mathbb{R}^{N \times s \times s}\) denote the attention scores of heads \(h_i\) and \(h_j\), respectively. Then \(D(P_{i} \parallel Q_{j})\) is calculated as:
\begin{equation}
\small
D(P_{i} \parallel Q_{j}) = \frac{1}{N \times s} \sum_{n=1}^{N \times s} \sqrt{\frac{1}{2} D_{KL}(P_{i}^{(n)} \parallel M_{ij}^{(n)}) + \frac{1}{2} D_{KL}(Q_{j}^{(n)} \parallel M_{ij}^{(n)})}
\end{equation}
where \(M_{ij}^{(n)} = \frac{1}{2}(P_{i}^{(n)} + Q_{j}^{(n)})\), \(D_{KL}\) denotes the Kullback-Leibler Divergence, and \(D_{ij}\) represents the average Jensen-Shannon divergence between heads \(h_i\) and \(h_j\) across the dataset. By visualizing the attention heads as graph nodes and connecting nodes with a divergence less than a predefined threshold \(\tau\) via an edge, we can clearly illustrate the relationships between these heads. Fig.~\ref{fig:3} demonstrates that some heads fall into the same group, signaling information redundancy, whereas others stand alone, highlighting the uniqueness of their information. We also observe that specific layers form large groups, indicating higher redundancy. The details of our pruning strategy for the attention module are outlined in Algo~\ref{algo:1}.

\begin{figure}
\begin{minipage}[t]{0.48\textwidth}
\begin{algorithm}[H]
\tiny
\caption{Attention Heads Pruning.}
\begin{algorithmic}[numbered=true]
\State \textbf{Input:} Pairwise head divergence matrix $D \in \mathbb{R}^{h \times h}$
\State \textbf{Input:} divergence threshold $\tau$
\State \textbf{Output:} List of candidate heads for pruning $C$
\State Initialize $C = []$
\For{row $i$ \textbf{and} col $j$ \textbf{in} $D$}
    \If{$D[i][j] < \tau$ \textbf{and} $(i, j) \notin edges$ \textbf{and} $(j, i) \notin edges$ \textbf{and} $i \neq j$}
        \If{$i \notin C$ \textbf{and} $j \notin C$}
            \State $C.append(i)$
        \EndIf
    \EndIf
\EndFor
\State \textbf{Prune} $C$
\end{algorithmic}
\label{algo:1}
\end{algorithm}
\end{minipage}%
\hfill
\begin{minipage}[t]{0.48\textwidth}
\begin{algorithm}[H]
\tiny
\caption{Pre-Pruning Recovery.}
\begin{algorithmic}[numbered=false]
\State \textbf{Input:} Depth-2 module $m_i$ with $W_1$ and $W_2$; Input $X$
\State \textbf{Input:} Original dense outputs $Y_1, Y_2$ for $X$
\State \textbf{Input:} Preceding pruned modules $m_1..m_{i-1}$
\State \textbf{Output:} Reconstructed weight $\bar{W}_{1}$ and $\bar{W}_{2}$
\Procedure{Weights Reconstruction}{}
% \State {\# pruned modules: \{$m_1..m_{i-1}$\}}
\State $\hat{X}_1 \gets (m_{i-1}(m_{i-2}..(m_{1}X)))$ \\
- - - - - - - - - - - - - - - - - - - - - - - - - - - - - - - -
\State $\bar{W}_{1} \gets (\hat{X}_{1}\hat{X}_{1}^{\top})^{-1}\hat{X}_{1}^{\top}Y_{1}$
\State $\hat{Y}_{1} \gets \bar{W}_{1}\hat{X}_{1}$ \\
- - - - - - - - - - - - - - - - - - - - - - - - - - - - - - - -
\State $\hat{X}_2 \gets $ post-process($\hat{Y}_{1}$)
\State $\bar{W}_{2} \gets (\hat{X}_{2}\hat{X}_{2}^{\top})^{-1}\hat{X}_{2}^{\top}Y_{2}$
\EndProcedure
\end{algorithmic}
\label{algo:2}
\end{algorithm}
\end{minipage}
\end{figure}

\subsubsection{Second-moment-based Metric for Depth-2 Module}~\label{sec:2nd-moment}

To prune the identified depth-2 module, we follow the structure mentioned in Sec~\ref{sec:structure}, namely pruning output channels in the first level and input channels in the second level. Since the pruned channel indexes from these two directions must match, we have to consider them together. 

This paper proposes a 2nd-moment-based pruning metric that is simple to calculate and incorporates information from multiple layers. %both output and input channel directions.%
To facilitate understanding, we use a basic feed-forward module as an example. Specifically, we describe the module as \(f(X)=B\sigma(AX)\), where \(\sigma\) is the activation function, \(A\) and \(B\) represent the upward and downward projection weights, respectively. Let \(X \sim \mathcal{N}(0, \Sigma_X)\) represent the input with zero means. Let \(A_i\) be any output channel vector of \(A\), then \(A_i^{\top}X \sim \mathcal{N}(0, A_i^\top\Sigma A_i)\). Let \(B_i\) be any input channel vector of \(B\), and \(B_{ij}\) is a scalar within this vector. Assuming a linear activation function, the second moment of \(Y_{ij} := B_{ij}\sigma(A_i^TX)\) can be derived as follows:
\begin{equation}
\small
E[Y_{ij}^2] = E[B_{ij}^2 \sigma(A_i^T X)^2] = B_{ij}^2 E[(A_i^T X)(X^T A_i)] = B_{ij}^2(A_i^T E[XX^T] A_i) = B_{ij}^2(A_i^T \Sigma_{X} A_i)
\end{equation}
where \(E[Y_{ij}^2]\) can determine the contribution of a single weight \(B_{ij}\) to the corresponding value of out channel \(B_i\). However, instead of summing \(E[Y_{ij}^2]\) along the input channels direction for all \(B_{ij}\), we sum along the output channels direction. In this way, we obtain a 2nd moment value corresponding to a specific index of the inner channel (output channel of A and input channel of B). We further embed \(j\) directly into the formula, then the importance score for inner channel $i$ is:
\begin{equation}
\small
\mathcal{M}_{i} = \sum_j E[Y_{ij}^2] = \|B_i\|_2^2 E[(A_i^T X)(X^T A_i)] = \|B_i\|_2^2 (A_i^T \Sigma_{X} A_i)
\end{equation}
When the activation function is ReLU, we add a coefficient of 1/2 to the above equation. For GeLU or SiLU, our observations indicate that only a small portion of values fall within the non-linear region; therefore, we treat them as equivalent to ReLU. This approach offers more valuable information from the covariance matrix compared to methods based on output energy (first moment)~\cite{hagiwara1993removal,hagiwara1994simple,hu2016network}. Unlike some statistical methods that require calibration datasets to collect feature values and then calculate statistical properties, our method can flexibly integrate information from both input and output channels, whereas those methods are limited to focusing only on output channels. When there is no way to estimate $\Sigma_X$, our method naturally degenerates to a function approximation method by assuming $\Sigma_X$ is the identity matrix. More details about calculating $\mathcal{M}_{i}$ for attention and the feed-forward modules can be found in the Appendix.

\subsection{Pre-Pruning Recovery Without Retraining~\label{sec:recovery}}

With the selection of the pruning structure and criteria, this paper proposes a module-wise pruning approach. Similar to layer-wise pruning, we prune these depth-2 modules sequentially. To prune one of them, we calculate importance scores for its inner channels based on the module’s structure, weights, and inputs. Notably, due to errors introduced by pruning preceding modules, the input to the current module inevitably deviates from its dense version. Consequently, even without pruning the current module, a discrepancy between its output and the original output is unavoidable. Recall that our design philosophy is to approximate the output as closely as possible. Thus, it is crucial to reconstruct the weights of the current module before pruning. This reconstruction ensures that the output of this module can still align as closely as possible with the original, even with the new input. This way, the pruning criteria for each channel can be optimally up-to-date. Inspired by~\cite{li2023towards}, this paper presents a pre-pruning recovery technique in Algo~\ref{algo:2}.

With this technique, we can mitigate the errors accumulated from previously pruned modules without requiring model retraining. Unlike previous work, which primarily focuses on a single layer, our approach targets more complex structures, including intricate layer dependencies. We provide more details on applying this method to the attention and feed-forward modules in the Appendix.

\section{Experiment}

This section initially presents the fundamental setup for our experiments. Subsequently, we demonstrate the results of experiments and provide an in-depth analysis from multiple perspectives.

\subsection{Setup~\label{section:5-1}}

\begin{table}[!t]
\tiny
\centering
\caption{The zero-shot performance of the compressed LLaMA-7B (20\% sparsity). Following the LLM-Pruner methodology~\cite{ma2023llm}, we only prune the transformer blocks from the 4th to the 30th. The average performance is calculated across seven classification datasets. 'Bold' indicates the best pruning-only performance, while 'underline' represents the overall best performance.}
\label{your-label}
\renewcommand{\arraystretch}{1.3}
\begin{tabular}{l|cc|ccccccc|c} % corrected number of columns
\hline
\rowcolor[HTML]{EFEFEF}
\textbf{Pruning Methods} & \textbf{WikiText2~$\downarrow$} & \textbf{PTB~$\downarrow$} & \textbf{BoolQ} & \textbf{PIQA} & \textbf{HellaSwag} & \textbf{WinoGrande} & \textbf{ARC-e} & \textbf{ARC-c} & \textbf{OBQA} & \textbf{Ave~$\uparrow$} \\ \hline
Dense~\cite{touvron2023llama,ma2023llm}                     & 12.62              & 22.14        & 73.18         & 78.35        & 72.99            & 67.01     & 67.45        & 41.38         & 42.4          & 63.5         \\ \hline
\rowcolor{blue!10}\multicolumn{11}{l}{Data Free Pruning} \\ \hline
Random~\cite{hoefler2021sparsity}            & 23.02             & 40.19       & 46.21         & 71.33        & 59.35            & 56.51             & 47.97         & 32.0          & 36.30          & 49.95 \\
L1 norm~\cite{hoefler2021sparsity}                 & 179.02             & 311.75       & 51.28         & 60.22        & 43.14            & 52.01             &     36.53& 27.89         & 30.8          & 43.12          \\
L2 norm~\cite{hoefler2021sparsity}                 & 582.41             & 1022.17      & 60.18         & 58.54        & 37.04            & 53.27             & 32.91        &       27.56 &    29.8          & 42.76      \\ \hline
\rowcolor{orange!10} Ours (Self-Gen)          & 21.76              & 34.3         & 63.51         & 72.63        & 56.54         &    54.46   & 51.68             & 33.79         & 36.4          & 52.72        \\
\rowcolor{orange!10} Ours SG w/ remedy  &  \textbf{20.32}     &  \textbf{33.42}     &     \textbf{64.17}    &  \textbf{72.67}     &  \textbf{58.43}       &      \textbf{57.29}      &   \textbf{53.32}        & \textbf{34.15}    &    \textbf{37.23}     &  \textbf{53.89}  \\ \hline
\rowcolor{blue!10}\multicolumn{11}{l}{Data Dependent Pruning} \\ \hline
\multicolumn{11}{c}{\textcolor{blue}{Training-Aware Pruning}} \\ \hline
LLM-Pruner Vec~\cite{ma2023llm}       & 22.28              & 41.78        & 61.44         & 71.71        & 57.27            & 54.22             & 55.77         & 33.96         &    38.4        & 53.52         \\
LLM-Pruner E1~\cite{ma2023llm} & 19.09          & 34.21        & 57.06         & 75.68        & 66.8             & 59.83             & 60.94         & 36.52         &  40.0     & 56.69          \\
LLM-Pruner E2~\cite{ma2023llm}     & 19.77              & 36.66        & 59.39         & 75.57        & 65.34            & 61.33             & 59.18         & 37.12         & 39.8      & 56.82       \\ \hline
\multicolumn{11}{c}{\textcolor{blue}{Inference-Aware Pruning}} \\ \hline
Wanda-sp~\cite{sun2023simple}    & 27.45              & 49.52        & 64.16         & 75.21        & \underline{\textbf{68.62}}            & 62.27             & 59.68         & 36.68         & 39.2        & 57.97        \\ \hline
\rowcolor{orange!10}Ours (Calibration)      & \textbf{17.48}              & \underline{\textbf{30.04}}        & 66.48         & 75.78        & 67.73            & 62.27             & 61.4          & 35.49         & 39.6        & 58.39        \\
\rowcolor{orange!10}Ours C w/ remedy      &  17.90             & 31.23        & \underline{\textbf{70.12}}         & \underline{\textbf{76.86}}       & {68.55}            & \underline{\textbf{65.76}}             & \underline{\textbf{64.23}}         & \underline{\textbf{38.54}}        & \underline{\textbf{40.5}}         &  \underline{\textbf{60.65}} \\ \hline
\multicolumn{11}{c}{\textcolor{blue}{Retraining-required Pruning}} \\ \hline
LLM-P. LoRA~\cite{ma2023llm}      & \underline{17.37}        & 30.39        & 69.54         & 76.44        & 68.11            & 65.11             & 63.43         & 37.88         & 40.0       & 60.07        \\ \hline
\end{tabular}
\label{tab:1}
\vspace{-0.1cm}
\end{table}

\textbf{Baselines:} This paper presents a comprehensive comparison of state-of-the-art pruning methods across multiple dimensions, aiming for fair evaluations and in-depth analyses to uncover the reasons behind the observed results. First, we compare our approach with data-free pruning methods, including random pruning and magnitude-based pruning (L1 and L2 norms)~\cite{hoefler2021sparsity}. Next, we evaluate our methods against data-dependent pruning techniques, encompassing training-aware, inference-aware, and retraining-required methods. In the training-aware category, we compare with various configurations of LLM-Pruner~\cite{ma2023llm}, such as Element1, Element2, and Vector-wise magnitude pruning. Within the inference-aware category, we compare with the structured version of Wanda~\cite{sun2023simple} and FLAP~\cite{an2024fluctuation}. Additionally, we extend our comparisons to include the LLM-Pruner method augmented with retraining. Such comprehensive evaluations will demonstrate the effectiveness of our pruning approach.%This thorough evaluation demonstrates the effectiveness of our pruning approach.

\textbf{Models:} 
Our primary experiments are categorized into two series based on the model scale: LLaMA-7B with 7 billion parameters and GPT-2 with 110 million parameters~\cite{radford2019language, touvron2023llama}. This aligns with our study's goal to assess pruning performance across different model sizes and ensure a thorough examination. Additionally, we extend our experiments to other models, including LLaMA-13B, 
% LLaMA2 variants at both 7B and 13B~\cite{touvron2023llama2}, 
Vicuna-7B~\cite{chiang2023vicuna}.
% , GPT-2 at 1.5 billion parameters~\cite{radford2019language}. 
This comprehensive selection allows us to explore a broader spectrum of capabilities and sizes, enhancing our understanding of how different architectures perform under various computational constraints. Additional experiment results can be found in the Appendix.

\textbf{Evaluation and Datasets:}
To evaluate performance, we adopt LLaMa's approach by conducting zero-shot task classification on a range of common sense reasoning datasets: BoolQ~\cite{clark2019boolq}, PIQA~\cite{bisk2020piqa}, HellaSwag~\cite{zellers2019hellaswag}, WinoGrande~\cite{sakaguchi2021winogrande}, ARC-easy~\cite{clark2018think}, ARC-challenge~\cite{clark2018think}, and OpenbookQA~\cite{mihaylov2018can}. Following the methodology in~\cite{gao2021framework}, the model either ranks the options in multiple-choice tasks or generates answers in open-ended formats. Additionally, we enhance our evaluation by conducting a zero-shot perplexity (PPL) analysis on WikiText2~\cite{merity2016pointer} and the Penn Treebank (PTB)~\cite{marcus1993building}.

\textbf{Implementation:}
During the pruning phase, we randomly select 16 samples from Wikitext2 and Bookcorpus, truncated to a sequence length of 128 for LLaMA-7B and 1024 for GPT-2. These samples serve as calibration data for pruning metric calculation and covariance matrix extraction, respectively. During the recovery phase, we sample an additional 1,024 examples from the downstream dataset to guide optimization in the data-dependent comparison experiments.
\subsection{Results and Analysis}

\begin{wraptable}{r}{0.41\textwidth}
    \centering
    \vspace{-0.5cm}
    \tiny
    \caption{Similarity-based analysis for LLaMA-7B attention heads pruning (all blocks) with different $\tau$. 'Bold' indicates the best performance.}
    \renewcommand{\arraystretch}{1.3}
    \begin{tabular}{l|c|c|c}
        \hline
        Methods & \#~pruned heads & Wiki2~$\downarrow$ & PTB~$\downarrow$ \\
        \hline
        Dense & 0  & 12.62 & 22.14 \\
        \hline
        \rowcolor{orange!10}Ours ($\tau=0.16$) & 88 & \textbf{12.96} & \textbf{22.45} \\
        Random & 64 & 14.50 & 24.13 \\
        L2 Norm & 64& 14.69 & 25.64 \\
        1st+2nd order & 64 & 13.45 & 24.19 \\
        FLAP & 88 & 12.90 & 22.67 \\ \hline
        \rowcolor{orange!10}Ours ($\tau=0.19$) & 204 & 14.69 & \textbf{24.32} \\
        Random & 192 & 18.75 & 35.73 \\ 
        L2 Norm & 192 &  195.84 & 371.65 \\
        1st+2nd order & 192 & 14.81 & 28.77 \\
        FLAP & 204 & \textbf{13.22} & 24.42 \\
        \hline
    \end{tabular}
    \vspace{-0.5cm}
    \label{tab:3}
\end{wraptable}

We present the main results in Tab.~\ref{tab:1}. For the data-free comparison experiments, we leverage the inherent ability of LLMs to generate sentences. Our pruning method uses these generated sentences as calibration data because, given that the LLMs are well-trained, these sentences naturally conform to the semantic and syntactic token distributions of the training data. Compared to traditional data-free metrics (L1 or L2), our data-free version, which relies solely on the model itself, achieves significant improvements in perplexity and up to a 20\% enhancement in zero-shot evaluation for downstream tasks. Moreover, our method surpasses random pruning by at least 6\%, a significant improvement achieved without relying on existing datasets, while traditional metrics (L1 or L2) fail to outperform. These results demonstrate the superiority of our techniques in data-free pruning methods.

\begin{wrapfigure}{l}{0.41\textwidth}
  \centering
  \vspace{-0.5cm}
  \includegraphics[trim=10 10 30 10, clip, width=0.40\textwidth]{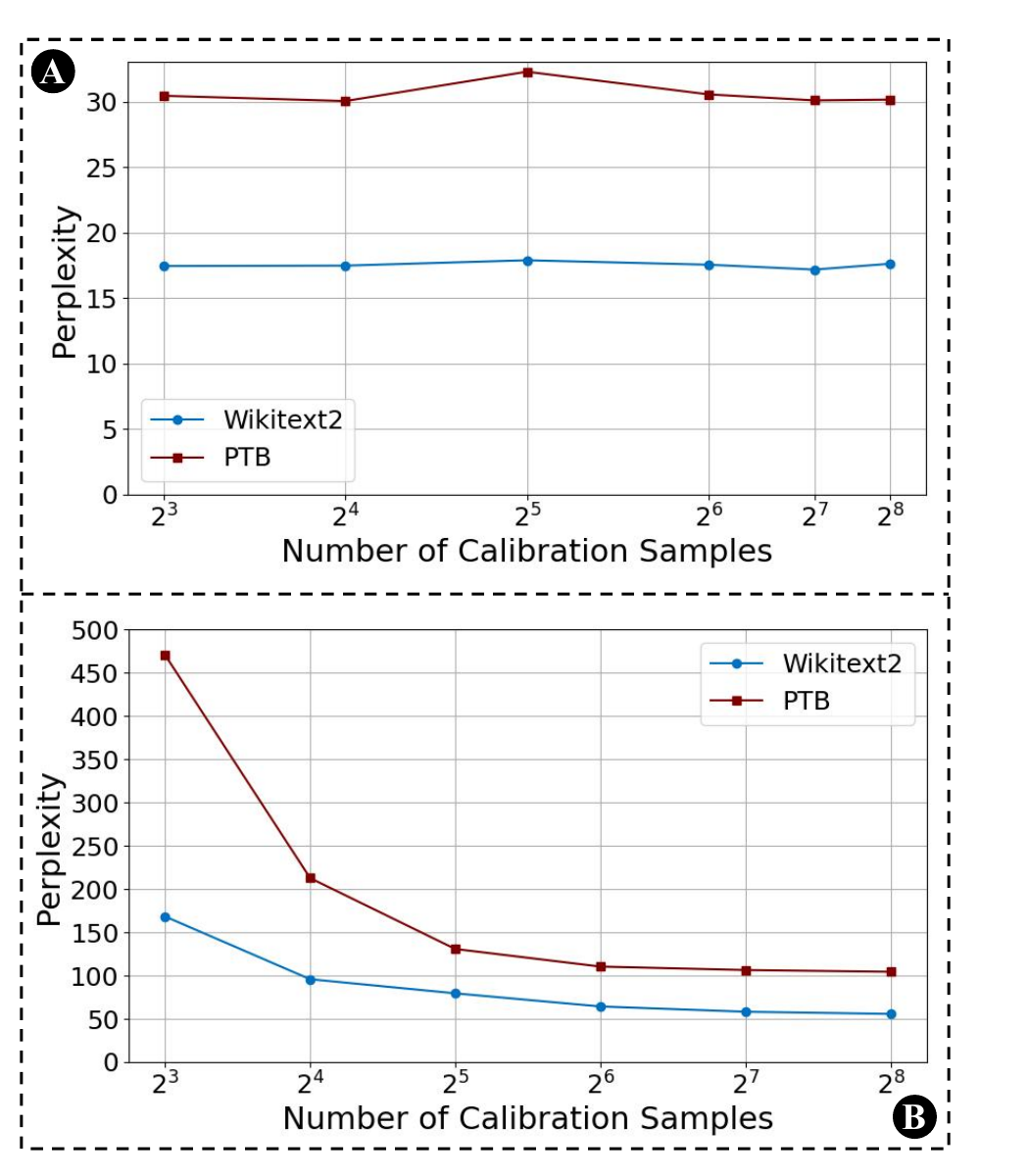}
  \captionof{figure}{Performance of compressed \textbf{A}: LLaMA-7B~(w/o Remediation) and \textbf{B}: GPT-2~(w/ Remediation) concerning the number of calibration samples.}
  \label{fig:4}
  \vspace{-1.5cm}
\end{wrapfigure}

Our approach outperforms data-dependent pruning methods and the inference-only method Wanda-SP. Impressively, it also surpasses the state-of-the-art training-aware pruning method LLM-Pruner, which includes different configurations such as Element1, Element2, and Vector. Our approach consistently demonstrates better pruning results without requiring computationally intensive first-order and second-order information. Moreover, our method even achieves better results compared to LLM-Pruner with LoRA, despite the latter involving model retraining.

We also compare our method with the state-of-the-art inference-only method FLAP and present the results in Tab.~\ref{tab:2}. Our approach exhibits significantly better results on the GPT-2 model and achieves comparable performance with LLaMA-7B. Overall, our method demonstrates superior performance in both data-free and data-dependent pruning categories.

\begin{table}[ht]
\tiny
\centering
\caption{Perplexity of compressed GPT-2 and LLaMA-7B (25\% and 50\% sparsity) on Wikitext2 and PTB. We prune the 4th to 30th transformer blocks for LLaMA-7B and all blocks for GPT-2. 'Bold' indicates the best performance, while 'underline' represents the second-best performance.}\
\renewcommand{\arraystretch}{1.3}
\begin{tabular}{p{3.6cm}|cc|cc|cc|cc}
\hline
\rowcolor[HTML]{EFEFEF}
\textbf{Models} & \multicolumn{4}{c|}{\textbf{GPT-2: [0-12)}} & \multicolumn{4}{c}{\textbf{LLama-7b: [4-30)}} \\ \hline
\rowcolor[HTML]{EFEFEF}
\textbf{Datasets:~PPL} & \multicolumn{2}{c|}{\textbf{WikiText2: 29.95}~\cite{radford2019language}~$\downarrow$} & \multicolumn{2}{c|}{\textbf{PTB: 40.12}~\cite{radford2019language}~$\downarrow$} & \multicolumn{2}{c|}{\textbf{WikiText2: 12.62}~\cite{ma2023llm}~$\downarrow$} & \multicolumn{2}{c}{\textbf{PTB: 22.14}~\cite{ma2023llm}~$\downarrow$} \\
\hhline{---------}
Sparsity & \textbf{25\%} & \textbf{50\%} & \textbf{25\%} & \textbf{50\%} & \textbf{25\%} & \textbf{50\%} & \textbf{25\%} & \textbf{50\%} \\
\hline
\rowcolor{blue!10}\multicolumn{9}{l}{Data Free Pruning} \\ \hline
Random~\cite{hoefler2021sparsity} & \underline{189.73} & 1839.33 & \underline{245.33} & 2769.6 & \underline{23.02} & \underline{100.42} & \underline{40.19} & \underline{133.56} \\
L1 norm~\cite{hoefler2021sparsity} & 338.3 & 1226.13 & 583.2 & 1290.45 & 179.02 & 891.23 & 311.75 & 1034.69 \\
L2 norm~\cite{hoefler2021sparsity} & 227.32 & \underline{674.52} & 324.33 & \underline{800.14} & 582.41 & 14000.68 & 1022.17 & 28062.45 \\
\rowcolor{orange!10}Ours (Self-Generation w/ remedy) & \textbf{119.29} & \textbf{586.87} & \textbf{152.93} & \textbf{723.39} & \textbf{21.76} & \textbf{58.61} & \textbf{34.3} & \textbf{64.24} \\\hline
\rowcolor{blue!10}\multicolumn{9}{l}{Data Dependent Pruning} \\ \hline
\multicolumn{9}{c}{\textcolor{blue}{Training-Aware Pruning}} \\ \hline
LLM-Pruner Element1~\cite{ma2023llm} & 9229.32 & 32453.23 & 11993.24 & 8020.87 & 19.09 & 48.84 & 34.21 & 105.24 \\
LLM-Pruner Element2~\cite{ma2023llm} & 1897.32 & 14706.23 & 2258.33 & 18598.33 & 19.77 & 72.89 & 36.66 & 138.33 \\
LLM-Pruner Vector~\cite{ma2023llm} & 488.32 & 39025.12 & 6169.56 & 18616.87 & 22.88 & 55.68 & 41.76 & 305.24 \\
\hline
\multicolumn{9}{c}{\textcolor{blue}{Inference-Aware Pruning}} \\ \hline
Wanda-Structured Pruning~\cite{sun2023simple} & 586.34 & 4147.32 & 355.17 & 3246.79 & 27.45 & 69.02 & 49.52 & 132.52 \\\hline
FLAP UL-UM w/o remed~\cite{an2024fluctuation} & 818.14 & 3636.23 & 554.32 & 2758.37 & 17.15 & 36.08 & 34.96 & 85.22 \\
FLAP UL-UM w/ remedy~\cite{an2024fluctuation} & 2197.32 & 3043.35 & 2199.24 & 3561.76 & \underline{15.76} & \underline{26.87} & 32.1 & 66.18 \\
\rowcolor{orange!10}Ours UL-UM (Calibration w/o remedy) & \underline{81.96} & \underline{317.37} & 186.68 & \underline{936.57} & NA & NA & NA & NA \\
\hline
FLAP AL-AM w/o remedy~\cite{an2024fluctuation} & 126.57 & 5538.32 & \underline{135.07} & 10244.95 & {17.01} & 34.09 & 30.99 & 71.76 \\
FLAP AL-AM w/ remedy~\cite{an2024fluctuation} & 1349.25 & 5382.14 & 1769.56 & 7476.08 & \textbf{15.06} & \textbf{26.55} & \textbf{29.45} & \underline{\textbf{57.89}} \\
\rowcolor{orange!10}Ours ML-MM (Calibration w/o remedy) & \textbf{79.4} & \textbf{251.34} & \textbf{130.54} & \textbf{756.33} & 17.48 & \underline{26.87} & \underline{30.04} & \underline{\textbf{57.89}} \\
\hline
\end{tabular}
\label{tab:2}
\vspace{-0.1cm}
\end{table}

\subsection{Ablation Study}

We also explore our pruning metrics by exclusively pruning attention heads. The experimental results in Tab.~\ref{tab:3} demonstrate that for colossal LLMs like LLaMA-7B, our similarity analysis effectively identifies redundant attention heads with minimal negative impact on model performance. Compared to inference-aware metrics such as the L2 norm, training-aware metrics using first- and second-order information, and random pruning, our similarity-based metric consistently outperforms. When compared to the specifically designed metric of FLAP, we achieve better or comparable performance. These results strongly indicate that we should prioritize pruning redundant information rather than heads with small importance scores.

Additionally, we designed experiments to explore the influence of the number of calibration samples. Figure \ref{fig:4} shows that in LLaMA-7B pruning-only experiments, our method is insensitive to the number of calibration samples, achieving comparable results with as few as 8 samples and as many as 128 samples. Conversely, in GPT-2 pruning with remediation experiments, performance improves with an increasing number of calibration samples. These findings demonstrate that our pruning method is robust regardless of the number of calibration samples, while our pre-pruning recovery method benefits from a higher number of calibration samples. However, this improvement gradually diminishes once the number of samples reaches a critical threshold.

\section{Discussion, Limitation, and Conclusion}

\textbf{Implicit Motivation and Call:} In the pre-deep learning era, various pruning metrics and structures were designed. For example, variance-based pruning and bias-based remedy methods similar to FLAP were proposed by researchers 30 years ago~\cite{engelbrecht1999variance,sietsma1988neural,engelbrecht1996sensitivity,thimm1995evaluating}. These early researchers already recognized that feature information is at least as crucial as model weights in constructing pruning criteria. In the early stages of deep learning (before 2022), many researchers found that multi-round model retraining could easily recover the lost performance induced by pruning, even when based solely on weight magnitudes. As a result, the importance of pruning metrics and structure design was often overlooked, with reliance placed on retraining to validate methods. However, this paradigm shifted after 2022, when colossal LLMs became mainstream in the community. Training such models is prohibitively expensive, making pruning that relies on multi-round retraining impractical. Although parameter-efficient training methods like LoRA can reduce costs, they still require rigorous data selection~\cite{hu2021lora,ma2023llm}. Thus, we urge the community to return to designing metrics that better account for the influence of both weights and features, rather than focusing solely on dataset competition. Motivated by this, this paper focuses on inference-aware pruning metrics that do 
not require retraining.

\textbf{Limitation:} This work evaluates the compressed LLMs primarily on perplexity and downstream tasks. However, we do not assess the emergent abilities of colossal LLMs, such as mathematical reasoning, safety alignment, common sense reasoning, contextual understanding, and creativity in text generation. Future research will focus on evaluating and enhancing these emergent abilities to provide a more comprehensive understanding of the compressed LLMs.

%\section*{Conclusion}
\textbf{Conclusion:} This paper introduces a novel approach to pruning large language models (LLMs) by identifying a depth-2 pruning structure and developing two inference-aware pruning criteria. These methods surpass traditional metrics and eliminate the need for computationally expensive retraining. Our two-step reconstruction technique further mitigates pruning errors, ensuring superior performance across various datasets and models. Overall, our approach significantly reduces computational costs and hardware requirements, offering an efficient solution for pruning colossal LLMs.

% \input{src/9-misc/9-ack}
% ack should only be included for camera-ready version

\newpage

\bibliographystyle{plainnat}
\bibliography{neurips_2024}

\newpage

\section{Appendix-A}
Pruning is a promising method that can effectively reduce model inference costs. In this paper, we discuss general pruning methods and various classification philosophies. We summarize previous work and categorize pruning from multiple perspectives: structured and unstructured, data-free and data-dependent, training-aware and inference-aware, and retraining-free and retraining-dependent. We also propose an innovative optimization-oriented view of pruning, which includes: \textbf{A:} Function Approximation, \textbf{B:} Output Approximation, and \textbf{C:} Objective Approximation. Our pruning pattern is designed based on this perspective.

Additionally, we review more aspects of pruning, including the most popular techniques in the pre-deep learning era (before 2022), such as \textbf{Iterative Magnitude Pruning} and the \textbf{comparison between random pruning and magnitude pruning}. We also discussed the \textbf{relationship between pruning and quantization}. By considering these various dimensions and methodologies, we aim to provide a comprehensive understanding of pruning and its potential to enhance model efficiency.

\subsection{Iterative Magnitude Pruning} 
Iterative Magnitude Pruning (IMP) is the most renowned strategy for achieving state-of-the-art results, surpassing other methods such as Single-shot Network Pruning (SNIP)~\cite{Frankle:2019, Frankle:2020, frantar2022optimal, Lee:2018}. This approach divides the pruning process into multiple stages by gradually increasing the sparsity. At each stage, the goal is to identify and remove redundant parameters or neurons. The most intuitive approach is to assign an importance score to each element and keep only the top-k elements, where the score can be based on the absolute value of weights, output sensitivity, gradients, or other fine-designed metrics~\cite{Hagiwara:1993, Gale:2019, Thimm:1995, Han:2015, Zhu:2017, Cuadros:2020, Luo:2017}. Weight magnitude is the most straightforward and data-free method, while other metrics can be computationally expensive as they require training with data~\cite{Yu:2018, He:2019, Golub:2019, Molchanov:2019, Singh:2020, Dong:2017}. Moreover, IMP is accompanied by a retraining phase to restore performance, which can be computationally costly. Therefore, in the era of colossal LLMs, IMP and other methods that heavily depend on model retraining are no longer effective due to the immense costs involved.

\subsection{Randomized Pruning v.s. Magnitude Pruning}

Excluding the influence of model retraining, we discovered an interesting phenomenon for model pruning. For colossal LLMs such as LLaMA-7B, randomized pruning surprisingly produced competitive results. Specifically, compared to traditional data-free pruning metrics like L1 and L2 norm values, randomized pruning achieved several times better results, even rivaling data-dependent pruning methods. However, this advantage only existed when the pruning ratio was less than 2x. As the pruning ratio increased, magnitude pruning gradually yielded better results. Initially, we attributed this phenomenon to the high redundancy of parameters in LLMs. However, our experiments with GPT-2 showed that randomized pruning was still weaker than magnitude pruning. \textbf{Therefore, we speculate that for colossal LLMs like Llama-7B, feature information plays a more crucial role in model activations compared to smaller LLMs like GPT-2.}

Magnitude-based pruning methods aim to remove weights or neurons from a neural network that appear least influential, primarily determined by the value of their weights. The rationale behind these methods is to reduce overall model size and computational requirements without a drastic loss in performance. However, several challenges arise with this approach, and one major challenge is the lack of variety if the magnitude is based on data-free metrics (L1 or L2). This kind of metric focuses solely on the magnitude of the weights for pruning decisions, potentially missing smaller weights that play pivotal roles, especially in edge cases or rarer instances.

To illustrate this more clearly, consider the following example. The output of a neural network can be represented as \( y = \sum(w_i \cdot f_i) \), where \( y \) is the network output, \( f_i \) represents a feature, and \( w_i \) is the corresponding weight. In magnitude-based pruning (L1 or L2), if \( |w_i| < \tau \) ($\tau$ is pruning threshold), then \( w_i \cdot f_i \) is pruned. However, the impact on \( y \) is not solely determined by \( w_i \), but by the combined effect of \( w_i \) and the sensitivity of \( f_i \). For instance, if \( f_i \) represents the sharpness of an image, even a small weight \( |w_i| = 0.01 \) can significantly affect \( y \) if \( f_i \) is highly sensitive, such as affecting object recognition. Conversely, if \( f_i \) represents the hue of an image background, a large weight \( w_i = 5 \) might have minimal impact on \( y \) if \( f_i \) is less sensitive, such as the background hue not altering recognition much. The influence on \( y \) is thus a joint effect of \( w_i \) and the sensitivity of \( f_i \). This example indicates that the influence of feature information plays a significant role in identifying redundant elements. 

\begin{wrapfigure}{l}{0.41\textwidth}
  \centering
  \includegraphics[trim=0 0 30 10, clip, width=0.40\textwidth]{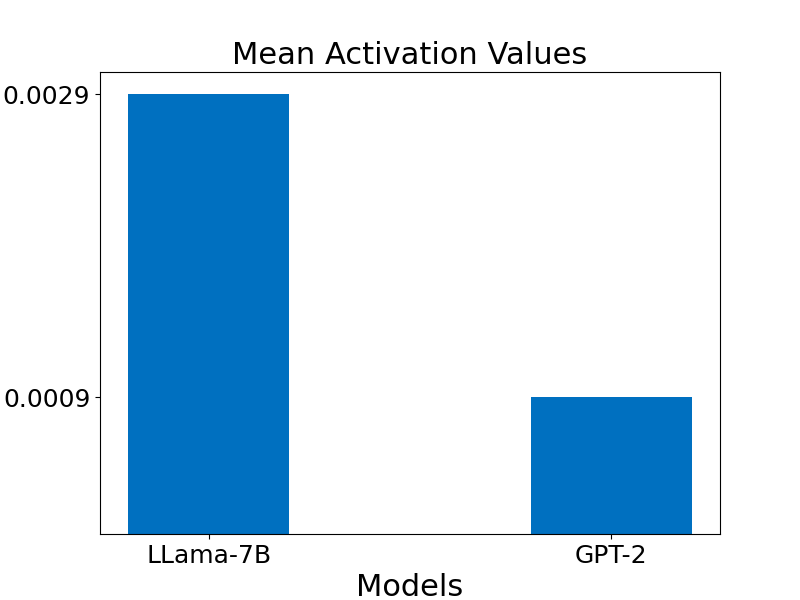}
  \captionof{figure}{Mean activation value of Llama-7B and GPT-2 on Wikitext2.}
  \label{fig:5}
  \vspace{-0.5cm}
\end{wrapfigure}

Based on the above observation, we speculate that LLaMA-7B's feature information contributes more to the importance score of removed elements when the pruning ratio is less than 2x. As the pruning ratio gradually increases, the influence of the features on the activation values is no longer greater than that of the weights. Therefore, randomized pruning fails at larger pruning ratios. To validate our hypothesis, we conducted a statistical analysis on the feature values of LLaMA-7B, described in Figure~\ref{fig:5}. Our results show that colossal LLMs like Llama-7B have larger activation values than smaller LLMs like GPT-2. \textbf{These findings further motivate us to design the pruning metrics that incorporate both feature and weight information instead of seeking dataset competition.}

\subsection{Pruning v.s. Quantization:} 

Pruning, though considered less effective than quantization in the era of colossal LLMs, should not be underestimated. In practice, pruning and quantization can complement each other, yielding significant benefits when applied together~\cite{frantar2022optimal}. Even pruning a small percentage of parameters, such as 5\%, can be valuable if it meets practical performance requirements. Therefore, integrating pruning into the optimization process is always worthwhile.

\section{Appendix-B}

In Section~\ref{sec:two-structure}, we introduced the 2nd moment-based pruning metric for a standard depth-2 module. However, there are different variants of depth-2 modules, including the attention module and the gated feed-forward module. We describe the calculation of the metric for these variants in the following section.

\textbf{Notations:} To better demonstrate our method, let us first establish the notations. We focus on the pruning of Transformer-based large language models, thus we refer to the attention mechanism as $\mathbf{Cat}_{i=1}^{h}[\sigma_{1}(\mathbf{X}\mathbf{W}_{i}^{K}\mathbf{W}_{i}^{Q}\mathbf{X}^\top)\mathbf{X}\mathbf{W}_{i}^{V}]\mathbf{W}^{O}$, with $i$ indicating the attention head index. The symbols $W^{K}$, $W^{Q}$, $W^{V}$ and $W^{O}$ represent the weights for the key, query, value, and output in the attention block, respectively. For the general and gated feed-forward module, we denote the logic as $\mathbf{W}^{D}\sigma_2(\mathbf{W}^{U}\mathbf{X})$ and $\mathbf{W}^{D}(\mathbf{W}^{U}\mathbf{X} \cdot \sigma_2(\mathbf{W}^{G}\mathbf{X}))$. Here, $W^{U}$, $W^{D}$, and $W^{G}$ stand for the weights for upward projection, downward projection, and gate projection. $\sigma$ refers to the activation function for all of them, which can be SoftMax, ReLU, GeLU, or SiLU function.

\subsection{Calculation of 2nd-Moment-Based Metric for Attention Module}

Based on the above notations, we can treat the entire output of $\sigma_{1}(\mathbf{X}\mathbf{W}{i}^{K}\mathbf{W}{i}^{Q}\mathbf{X}^\top)\mathbf{X}$ as the input to $\mathbf{W}_{i}^{V}$. Let $\hat{X}$ represent this new input. In this way, the attention module can be viewed as a module similar to a standard depth-2 module $(\hat{X}\mathbf{W}_{i}^{V})\mathbf{W}_{i}^{O}$, with each level having only one linear layer. However, we need to view an attention module as $m$ standard depth-2 modules ($m$ is the number of attention heads), as the attention heads operate independently.

\subsection{Calculation of 2nd-Moment-Based Metric for Gate Feed-forward Module}

For the gated feedforward module $\mathbf{W}^{D}(\mathbf{W}^{U}\mathbf{X} \cdot \sigma_2(\mathbf{W}^{G}\mathbf{X}))$, we treat it as a product of two standard depth-2 modules. Specifically, we can divide it into two modules: $\mathbf{W}^{D}\cdot\mathbf{W}^{U}\mathbf{X}$ and $\mathbf{W}^{D}(\sigma_2(\mathbf{W}^{G}\mathbf{X}))$, and calculate the 2nd-moment metric separately for them. Finally, we use the product of their own metric as the 2nd-moment metric for the entire module. 

Overall, these approaches allow us to effectively prune channels in both attention and gated feedforward modules by leveraging the 2nd moment-based metric.

\begin{algorithm}[H]
\small
\caption{Pre-Pruning Recovery for Attention Module.}
\begin{algorithmic}[1]
\State \textbf{Input:} Attention ModuleModel layers with weights $\{k\_proj, q\_proj, v\_proj, o\_proj\}$
\State \textbf{Input:} Corresponding inputs $x$
\State \textbf{Input:} Corresponding outputs $y\_k\_proj, y\_q\_proj, y\_v\_proj, y\_o\_proj$
\State \textbf{Output:} Reconstructed weights $\{\bar{W}_{k\_proj}, \bar{W}_{q\_proj}, \bar{W}_{v\_proj}, \bar{W}_{o\_proj}\}$
\Procedure{Weights Reconstruction}{}
    \State $xtx \gets \text{Matmul}(x.T, x)$
    \State $\bar{W}_{k\_proj} \gets \text{reconstruct\_best\_weight}(\text{xtx}, x, y\_k\_proj, w=W_{k\_proj})$
    \State $\bar{W}_{q\_proj} \gets \text{reconstruct\_best\_weight}(\text{xtx}, x, y\_q\_proj, w=W_{q\_proj}))$
    \State $\bar{W}_{v\_proj} \gets \text{reconstruct\_best\_weight}(\text{xtx}, x, y\_v\_proj, w=W_{v\_proj}))$
    \State $\text{restore\_layer\_weights}(\text{module.k\_proj}, \bar{W}_{k\_proj})$
    \State $\text{restore\_layer\_weights}(\text{module.q\_proj}, \bar{W}_{q\_proj})$
    \State $\text{restore\_layer\_weights}(\text{module.v\_proj}, \bar{W}_{v\_proj})$
    \State \# - - - - - - - - - - - - - - - - - - - - - - - - - - - - - - - -
    \State $x_2 \gets \text{module}(x)$ \# new input for o proj
    \State $xtx_2 \gets \text{Matmul}(x_2.T, x_2)$
    \State $\bar{W}_{o\_proj} \gets \text{reconstruct\_best\_weight}(xtx_2, x_2, y\_o\_proj, w=w=W_{o\_proj})$
    \State $\text{restore\_layer\_weights}(\text{module.o\_proj}, \bar{W}_{o\_proj})$
\EndProcedure
\end{algorithmic}
\label{algo:3}
\end{algorithm}

\section{Appendix-C}

In Section~\ref{sec:recovery}, we only provide the pre-pruning recovery algorithm for the standard depth-2 module, thus we describe the details of the recovery process for the attention module and gated feed-forward module in Algo~\ref{algo:3} and Algo~\ref{algo:4}, respectively.

\begin{algorithm}[H]
\small
\caption{Pre-Pruning Recovery for Gate Feed-forward Module.}
\begin{algorithmic}[1]
\State \textbf{Input:} Gated Feed-forward Module with weights $\{up\_proj, gate\_proj, down\_proj\}$
\State \textbf{Input:} Corresponding inputs $x$
\State \textbf{Input:} Corresponding outputs $y\_up\_proj, y\_gate\_proj, y\_down\_proj$
\State \textbf{Output:} Reconstructed weights $\{\bar{W}_{up\_proj}, \bar{W}_{gate\_proj}, \bar{W}_{down\_proj}\}$
\Procedure{Weights Reconstruction}{}
    \State $xtx \gets \text{Matmul}(x.T, x)$
    \State $\bar{W}_{up\_proj} \gets \text{reconstruct\_best\_weight}(\text{xtx}, x, y\_up\_proj, w=W_{up\_proj})$
    \State $\bar{W}_{gate\_proj} \gets \text{reconstruct\_best\_weight}(\text{xtx}, x, y\_gate\_proj, w=W_{gate\_proj}))$
    \State $\text{restore\_layer\_weights}(\text{module.up\_proj}, \bar{W}_{up\_proj})$
    \State $\text{restore\_layer\_weights}(\text{module.gate\_proj}, \bar{W}_{gate\_proj})$
    \State \# - - - - - - - - - - - - - - - - - - - - - - - - - - - - - - - -
    \State $x_2 \gets \text{module}(x)$ \# new input for down proj
    \State $ xtx_2 \gets \text{Matmul}(x_2.T, x_2)$
    \State $\bar{W}_{down\_proj} \gets \text{reconstruct\_best\_weight}(xtx_2, x_2, y\_down\_proj, w=W_{down\_proj})$
    \State $\text{restore\_layer\_weights}(\text{module.down\_proj}, \bar{W}_{down\_proj})$
\EndProcedure
\end{algorithmic}
\label{algo:4}
\end{algorithm}

\section{Appendix-D: Broader Impact}

Our proposed method efficiently prunes large language models with billions of parameters. Our proposal intends to mitigate AI risks from critical perspectives like economic inequality and concentration of power and further democratize the use of AI models. 

\begin{table}[!htb]
\tiny
\centering
\caption{The zero-shot performance of the compressed Vicuna-7B (20\% sparsity). Following the LLM-Pruner methodology~\cite{ma2023llm}, we only prune the transformer blocks from the 4th to the 30th. The average performance is calculated across seven classification datasets. 'Bold' indicates the best pruning-only performance, while 'underline' represents the overall best performance.}
\renewcommand{\arraystretch}{1.3}
\begin{tabular}{l|cc|ccccccc|c} % corrected number of columns
\hline
\rowcolor[HTML]{EFEFEF}
\textbf{Pruning Methods} & \textbf{WikiText2~$\downarrow$} & \textbf{PTB~$\downarrow$} & \textbf{BoolQ} & \textbf{PIQA} & \textbf{HellaSwag} & \textbf{WinoGrande} & \textbf{ARC-e} & \textbf{ARC-c} & \textbf{OBQA} & \textbf{Ave~$\uparrow$} \\ \hline
Dense~\cite{touvron2023llama,ma2023llm}                     & 16.11              & 61.37        & 76.57         & 77.75        & 70.64            & 67.40     & 65.11        & 41.21        & 40.80          & 62.78         \\ \hline
\rowcolor{blue!10}\multicolumn{11}{l}{Data Free Pruning} \\ \hline
Random~\cite{hoefler2021sparsity}            & 34.63             & 112.44       & 61.47         & 70.89        & 54.67            & 56.27             & 55.60         & 31.74          & 34.60          & 52.18 \\
L2 norm~\cite{hoefler2021sparsity}                 & 3339.98             & 5882.21      & 55.90         & 56.15        & 32.37            & 51.85             & 30.01        &       28.41 &    28.20          & 40.41      \\ \hline
\rowcolor{orange!10} Ours SG w/o remedy         & \textbf{28.45}              & \textbf{92.3}         & \textbf{62.51}         & \textbf{72.63}        & \textbf{56.54}         &    \textbf{57.46}   & \textbf{58.68}             & \textbf{33.29}         & \textbf{36.2}          & \textbf{53.91}        \\
\rowcolor{blue!10}\multicolumn{11}{l}{Data Dependent Pruning} \\ \hline
LLM-Pruner Vec~\cite{ma2023llm}       & 27.03              & 92.51        & 62.17         & 71.44        & 55.80            & 53.43             & 55.77         & 33.28        &    37.80        & 52.81         \\
LLM-Pruner E2~\cite{ma2023llm} & 24.70          & 94.34        & 62.87         & 75.41        & 64.00             & 58.41             & 60.98         & 37.12         &  39.00     & 56.83          \\
LLM-Pruner E1~\cite{ma2023llm}     & 25.74              & 92.88        & 61.70         & 75.30        & 63.75            & 56.20             & 63.22        & 36.60        & 37.00      & 56.25       \\
\rowcolor{orange!10}Ours (C) w/o remedy      & \textbf{19.88}              & \textbf{90.04}        & 62.48         & 75.68        & 65.23            & 61.27             & 63.4          & 35.49         & 37.6        & \textbf{57.31}        \\\hline
\rowcolor{blue!10}\multicolumn{11}{l}{Data Dependent Pruning w/ Retraining} \\ \hline
LLM-Pruner LoRA~\cite{ma2023llm}     & \underline{18.97}              & \underline{76.78}        & 60.40         & 75.63        & 65.45            & 63.22             & 63.05        & 37.71        & 39.00      & \underline{57.78}       \\ \hline
\end{tabular}
\label{tab:4}
\vspace{-0.1cm}
\end{table}
\begin{table}[!htb]
\tiny
\centering
\caption{The zero-shot performance of the compressed Llama-13B (20\% sparsity). The average performance is calculated across seven classification datasets. 'Bold' indicates the best pruning-only performance, while 'underline' represents the overall best performance.}
\renewcommand{\arraystretch}{1.3}
\begin{tabular}{l|cc|ccccccc|c} % corrected number of columns
\hline
\rowcolor[HTML]{EFEFEF}
\textbf{Pruning Methods} & \textbf{WikiText2~$\downarrow$} & \textbf{PTB~$\downarrow$} & \textbf{BoolQ} & \textbf{PIQA} & \textbf{HellaSwag} & \textbf{WinoGrande} & \textbf{ARC-e} & \textbf{ARC-c} & \textbf{OBQA} & \textbf{Ave~$\uparrow$} \\ \hline
Dense~\cite{touvron2023llama,ma2023llm}                     & 11.58              & 20.24        & 68.47         & 78.89        & 76.24            & 70.09     & 74.58        & 44.54        & 42.00          & 64.97         \\ \hline
\rowcolor{blue!10}\multicolumn{11}{l}{Data Free Pruning} \\ \hline
Random~\cite{hoefler2021sparsity}            & 19.24             & 31.84       & 63.33         & 73.18        & 63.54            & 60.85             & 64.44         & 36.26          & 38.00          & 57.09 \\
L2 norm~\cite{hoefler2021sparsity}                 & 61.15             & 91.43      & 61.50         & 67.57        & 52.90            & 57.54             & 50.13        &       31.14 &    36.80          & 51.08      \\ \hline
\rowcolor{orange!10} Ours SG w/o remedy          & \textbf{18.47}              & \textbf{29.87}         & 66.51         & 74.63        & 68.54         &    61.35   & 66.80             & 36.26         & 38.41          & \textbf{58.92}       \\
\rowcolor{blue!10}\multicolumn{11}{l}{Data Dependent Pruning} \\ \hline
LLM-Pruner Channel~\cite{ma2023llm} & 49.03          & 106.48        & 62.39         & 66.87        & 49.17            & 58.96            & 49.62         & 31.83         &  33.20     & 50.29          \\
LLM-Pruner E1~\cite{ma2023llm}     & 16.01              & 29.28      & 67.68        & 77.15        & 73.41            & 65.11             & 68.35        & 38.40        & 42.40      & 61.79       \\
\rowcolor{orange!10}Ours (C) w/o remedy      & \textbf{15.90}              & \textbf{28.33}       & 68.48         & 77.78        & 74.73            & 65.01             & 68.90          & 39.40         & 43.11        & \textbf{62.48}        \\\hline
\rowcolor{blue!10}\multicolumn{11}{l}{Data Dependent Pruning w/ Retraining} \\ \hline
LLM-Pruner LoRA~\cite{ma2023llm}     & \underline{15.18}              & \underline{28.08}       & 70.31         & 77.91        & 75.16            & 67.88             & 71.09        & 42.41        & 43.40      & \underline{64.02}       \\ \hline
\end{tabular}
\label{tab:5}
\vspace{-0.1cm}
\end{table}

%%%%%%%%%%%%%%%%%%%%%%%%%%%%%%%%%%%%%%%%%%%%%%%%%%%%%%%%%%%%

\end{document}